\pdfoutput=1
\documentclass[11pt]{article}

\usepackage{EMNLP2023}

\usepackage{times}
\usepackage{latexsym}
\RequirePackage[inline,shortlabels]{enumitem}
\setlist{nosep}
\usepackage[T1]{fontenc}
\usepackage[utf8]{inputenc}

\usepackage{microtype}

\usepackage{inconsolata}

\usepackage{breqn}
\usepackage{rotating}
\usepackage{float}
\usepackage{xcolor}
\usepackage{algorithm2e}
\usepackage{caption}
\usepackage{subcaption}
\usepackage{tabu}
\usepackage{multirow}
\usepackage{longtable}
\usepackage{array}
\usepackage{pifont}
\usepackage{hhline}
\usepackage{textcomp}
\usepackage{makecell}
\usepackage{amsmath}
\usepackage{amssymb}
\usepackage{graphicx}
\usepackage{adjustbox}
\usepackage{float}
\usepackage{notoccite}
\usepackage{booktabs}
\usepackage{defs,defsCustom}
\usepackage{todonotes}
\usepackage{enumitem}
\usepackage{listings}
\usepackage{pgfplots}
\usepackage{algorithm2e}
\usepackage{soul}
\usepackage{xcolor}

\usepgfplotslibrary{groupplots}
\usepackage{tikzscale}

\newcommand\blfootnote[1]{%
  \begingroup
  \renewcommand\thefootnote{}\footnote{#1}%
  \addtocounter{footnote}{-1}%
  \endgroup
}

\newcommand{\Any}{EitherInTopK}
\newcommand{\All}{BothInTopK}

\newcommand{\benchName}{AmbiQT}
\newcommand{\algName}{LogicalBeam}

\pgfplotsset{compat=newest, nodes near coords, nodes near coords style={font=\bfseries}}

\title{Benchmarking and Improving Text-to-SQL Generation under Ambiguity}
\author{Adithya Bhaskar$^{*\spadesuit\diamondsuit}$ \: Tushar Tomar$^{*\spadesuit}$ \: Ashutosh Sathe$^{\spadesuit}$ \: Sunita Sarawagi$^{\spadesuit}$\\$^{\spadesuit}$IIT Bombay \: $^{\diamondsuit}$Princeton University}

\newlength{\mylength}
\makeatletter
\newcommand{\mycfs}[1]{%
  \normalsize
  \@defaultunits\mylength=#1pt\relax\@nnil
  \edef\@tempa{{\strip@pt\mylength}}%
  \ifx\protect\@typeset@protect
     \edef\@currsize{\noexpand\mycfs\@tempa}% store calculated size
  \fi
  \mylength=1.2\mylength
  \edef\@tempa{\@tempa{\strip@pt\mylength}}%
  \expandafter\fontsize\@tempa
  \selectfont
}
\makeatother
\begin{document}
\maketitle
\begin{abstract}
Research in Text-to-SQL conversion has been largely benchmarked against datasets where each text query corresponds to one correct SQL.  However, natural language queries over real-life databases frequently involve significant ambiguity about the intended SQL due to overlapping schema names and multiple confusing relationship paths. To bridge this gap, we develop a novel benchmark called \benchName\ with over 3000 examples where each text is interpretable as two plausible SQLs due to lexical and/or structural ambiguity. 

When faced with ambiguity, an ideal top-$k$ decoder should generate all valid interpretations for possible disambiguation by the user~\cite{NL-Edit,zhong2022active}. We evaluate several Text-to-SQL systems and decoding algorithms, including those employing state-of-the-art LLMs, and find them to be far from this ideal. The primary reason is that the prevalent beam search algorithm and its variants, treat SQL queries as a string and produce unhelpful token-level diversity in the top-$k$.

We propose \algName, a new decoding algorithm that navigates the SQL logic space using a blend of plan-based template generation and constrained infilling. Counterfactually generated plans diversify templates while in-filling with a beam-search, that branches solely on schema names, provides value diversity. \algName\ is up to $2.5 \times$ more effective than state-of-the-art models at generating all candidate SQLs in the top-$k$ ranked outputs. It also enhances the top-$5$ Exact and Execution Match Accuracies on SPIDER and Kaggle DBQA\footnote{We release \benchName\ and \algName's implementation publicly at \url{https://github.com/testzer0/AmbiQT}.}.

\end{abstract}

\section{Introduction}
\blfootnote{$^*$  Equal Contribution. Work done while AB was at IIT Bombay. Correspondence to: <\texttt{adithyabcse@gmail.com}>}
Research on \texttosql\ generation has focused on scenarios where each natural language question is associated with one correct SQL~\cite{Mooney1, Mooney2, DUOrat, RATSQL, rubin-berant-2021-smbop-semi,  UnifiedSKG, PreT1, PreT2, PICARD, DinSQL}. Popular benchmarks driving such research, including  WikiSQL~\cite{WikiSQL}, SPIDER~\cite{SPIDER}, its robust perturbations~\cite{DrSpider}, and even ``in-the-wild'' benchmarks such as KaggleDBQA~\cite{kaggleDBQA} and SEDE~\cite{hazoom-etal-2021-text} all associate one correct SQL with text. Meanwhile, ambiguity is prevalent in real-life databases --- particularly the ones obtained by integrating several data sources for data analysis, where a natural language interface is most in demand.  The sources of ambiguity are several --- inherent ambiguity of natural language, the user's ignorance of table/column names, overlapping strings in column names, under-specified clauses, and confusion about whether aggregates are pre-computed, or if a join is required. \citet{hazoom-etal-2021-text} observe that up to 87\% of queries on the stack exchange database are under-specified, and \citet{ambiguous} mention that 11\% of queries exhibited ambiguity in column names. Although prior work has brought up ambiguity, there is no publicly available benchmark with ambiguous queries, nor a comprehensive evaluation of systems under ambiguity. 

Our first contribution is to bridge this gulf by developing a benchmark, \benchName, that tests \emph{performance under ambiguity} in the context of current models. \benchName\ includes over $3000$ examples, each associating a natural question on a database with \emph{two} valid SQLs. Inspired by our experience with several real-world datasets, we target four types of ambiguity spanning both lexical (ambiguous column and table names) and structural (whether a join is necessary, and an aggregate is pre-computed) ambiguity.
The benchmark is generated via a combination of ChatGPT \cite{ChatGPT} based synonym generation and perturbation, and standard rule-based perturbation.

When faced with ambiguity, an ideal Text-to-SQL system should incorporate all valid alternatives in their top-$k$ SQL outputs, for user resolution. We show that present approaches, ranging from T5-3B \cite{T5} to SOTA models, fail to generate all ambiguous outputs with any decoding strategy, including beam search and diversity-promoting sampling methods such as Nucleus~\cite{NucleusSampling} and Typical sampling~\cite{TypicalSampling}.  Most outputs are small lexical tweaks of the top choice, bringing about little meaningful diversity in SQL structures or schema alternatives. Even SOTA LLMs like ChatGPT \cite{ChatGPT} suffer from this issue.

To remedy the lack of diversity, we propose a new decoding algorithm, \algName, that allocates branching to explore underlying logical variants of the SQL rather than the string form. We catalog the errors of T5-3B \cite{T5} on the SPIDER \texttt{dev} split and use our insights to encourage targeted types of diversity --- the number of \texttt{JOIN}s and selections, and table/column names.

Our main contributions are:
\begin{itemize}[partopsep=-1pt,topsep=-2pt,parsep=-2pt,leftmargin=*, itemsep=0.4em]
    \item We develop \benchName, the first benchmark that tests performance under four types of ambiguity over \textbf{3000+} examples.
    \item We show that SOTA methods, including a fine-tuned T5-3B, RESDSQL~\cite{resdsql}, OpenAI Codex, and ChatGPT, provide a poor representation of ambiguity despite their high accuracy on conventional benchmarks.
    \item We present \algName, a two-step algorithm that generates plan-based templates with counterfactually controlled plan diversity and fills them via a beam search that branches only on schema names.
    \item We show that \algName\ consistently increases the fraction of time when all gold SQLs get generated in the Top-5 choices by $1.5-2.5\times$ over the baselines across the board on \benchName.
\end{itemize}

\section{Background and Related Work}
\begin{figure}[t]
    \centering
    \includegraphics[width=0.5\textwidth]{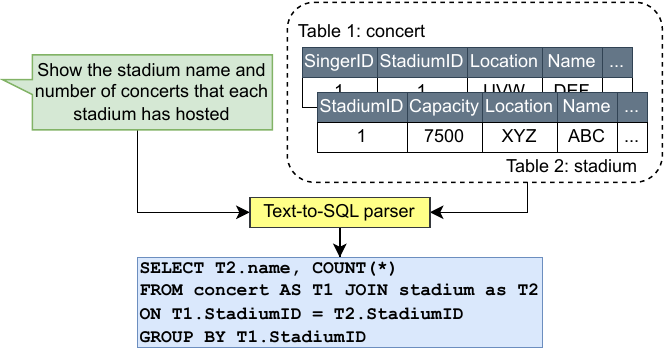}
    \caption{A Text-to-SQL system converts a user question to an SQL query, conditioned on the database schema and/or content.}
    \label{fig:text2SQL}
\end{figure}

%% Table AmbiSQL
\begin{table*}[t]
\setlength\tabcolsep{2.0pt}
    \centering
    \begin{small}
    \begin{tabular}{c|r|c|c|c}
    \toprule
    \multirow{2}{*}{\makecell{Kind of\\ambiguity}} & \multirow{2}{*}{Count} & \multicolumn{3}{c}{Example}\\
    \cline{3-5}
    \noalign{\vspace{0.3em}}
    & & Question Text & SQL \#1 & SQL \#2\\
    \hline
    \noalign{\vspace{0.3em}}
    \makecell{Column\\Ambiguity (C)} & 1240 & \makecell[l]{List the ids of\\ all students.} & \texttt{\makecell[l]{SELECT roll\_number\\FROM students}} & \texttt{\makecell[l]{SELECT admission\_number \\FROM students}}\\ 
    \hline
    \noalign{\vspace{0.3em}}
    \makecell{Table\\Ambiguity (T)} & 1417 & \makecell[l]{How many singers\\ do we have?} & \texttt{\makecell[l]{SELECT COUNT(*)\\FROM artist}} & \texttt{\makecell[l]{SELECT COUNT(*) FROM\\performer}}\\ 
    \hline
    \noalign{\vspace{0.3em}}
    \makecell{Join\\Ambiguity (J)} & 288 & \makecell[l]{What are the makers\\ and models?} & \texttt{\makecell[l]{SELECT maker, model\\FROM model}} & \texttt{\makecell[l]{SELECT t2.maker, t1.model FROM \\ model AS t1 JOIN model\_maker \\ AS t2 ON t1.model\_id = t2.model\_id}} \\ 
    \hline
    \noalign{\vspace{0.3em}}
    \makecell{Precomputed\\Aggregates (P)} & 101 & \makecell[l]{Find the average weight\\for each pet type.} & \texttt{\makecell[l]{SELECT AVG(weight), pettype\\ FROM pets GROUP BY pettype}} & \texttt{\makecell[l]{SELECT avg_weight, pettype\\FROM pets_weight}}\\
    \bottomrule
    \end{tabular}
    \end{small}
    \caption{The \benchName\ benchmark.  For each question, we list two valid SQL queries as per the schema.  The schema is not shown here, but the ambiguity in it can be inferred based on the two SQL queries.}
    \label{tab:BASE}
\end{table*}
A \texttosql\ model takes as input a question expressed as a natural language text $\vx$, and a database schema $\schema$ comprising of table and column names, and outputs an SQL program $\vy$ which can be executed against the database to answer the user's question. \fref{fig:text2SQL} shows an example. The training data for the task comprises  (text, schema, SQL) triplets spanning multiple distinct databases. 

\xhdr{Benchmarks} Popular benchmarks for the \texttosql\ task are WikiSQL~\cite{WikiSQL} and SPIDER~\cite{SPIDER}.  A few others have been proposed recently to capture real-world scenarios, such as KaggleDBQA~\cite{kaggleDBQA}, SEDE~\cite{hazoom-etal-2021-text}, and EHRSQL~\cite{EHRSQL}.  They all attach one SQL per text, though some of them mention the problem of ambiguity in real-world datasets.  Dr. SPIDER \cite{DrSpider}, designed to test the robustness of existing models, perturbs either the text or schema of SPIDER but still assigns one SQL per text.  

\paragraph{Ambiguity in SQL} Although ambiguity has been studied in other fields of NLP \cite{ambiguity1, ambiguity2, ambiguity3}, it has been unexplored in the context of semantic parsing. Ambiguity in SQL arising from related column names is discussed in \cite{ambiguous}, but they only consider column ambiguity. Their method of recognizing ambiguous queries depends on labeling words of the text and does not generalize to other kinds of ambiguity. To the best of our discernment, \benchName\ represents the first open benchmark for testing coverage of ambiguous alternatives.

\xhdr{Diverse Decoding} Prior work has critiqued the lack of meaningful diversity in beam-search outputs~\cite{BSNotGood1, BSNotGood2, BSNotGood3, BSNotGood4}. In response, many fixes have been proposed. Some proposals attempt to restrict the tokens sampled, using strategies like Nucleus sampling~\cite{NucleusSampling}, Truncated Sampling~\cite{hewitt-etal-2022-truncation}, and Typical Sampling~\cite{TypicalSampling}, while some rely on Template-Based decoding~\cite{Template1, Template2, CatSQL,TemplateS2T,ReFill}. A third approach is to generate a prefix with high diversity first, then generate the rest of the sentence with lower diversity. \citet{PlanGenerate} follow this recipe but focus on incorporating diverse entity orders in text summarization. 

\section{\benchName: A Benchmark of Ambiguous Text-to-SQL Conversion}

%% BS errors
\begin{figure*}[t]
{
\mycfs{8}
\centering
\begin{tabular}{p{0.9\textwidth}}
\toprule
\textbf{Question:} Show the names of high school students and their corresponding number of friends.\\
\midrule
\textbf{Gold Queries}\\
\begin{enumerate}[partopsep=-3pt,topsep=-2pt,parsep=-3pt,leftmargin=*, itemsep=0.4em]
\item SELECT t2.full\_name, count(*) FROM friend AS t1 JOIN highschooler AS t2 on t1.id = t2.id GROUP BY t1.id
\item SELECT t2.given\_name, count(*) FROM friend AS t1 JOIN highschooler AS t2 on t1.id = t2.id GROUP BY t1.id
\end{enumerate}\vspace*{-0.65em}\\
\midrule
\textbf{Outputs of T5-3B with Beam Search}\\
\begin{enumerate}[partopsep=-3pt,topsep=-2pt,parsep=-3pt,leftmargin=*, itemsep=0.4em]
\item \textcolor{green!60!black}{SELECT t1.given\_name, count(*) FROM highschooler AS t1 JOIN friend AS t2 on t1.id = t2.id GROUP BY t1.id}
\item SELECT t1.given\_name, count(*) FROM highschooler AS t1 JOIN friend AS t2 on t1.id = t2.id \textcolor{red!100}{GROUP BY t1.highschooler}
\item SELECT t1.given\_name, count(*) FROM highschooler AS t1 JOIN friend AS t2 on t1.id = t2.friend\_id \textcolor{red!100}{GROUP BY t1.grade}
\item SELECT \textcolor{orange!80!red}{t1.name}, count(*) FROM highschooler AS t1 JOIN friend AS t2 on t1.id = t2.id GROUP BY t1.id
\item SELECT \textcolor{orange!80!red}{t1.giving\_name}, count(*) FROM highschooler AS t1 JOIN friend AS t2 on t1.id = t2.id GROUP BY t1.id
\end{enumerate}\vspace*{-0.65em}\\
\bottomrule
\end{tabular}
\caption{Beam Search works \textcolor{green!60!black}{well} when targeting only one output, but leads to superficial diversity, for example via \textcolor{red!100}{different grouping} and \textcolor{orange!80!red}{erroneous variants} of column names.}
\label{fig:bserrors}
}
\end{figure*}

\benchName\ is constructed so that each text query has two distinct valid SQL interpretations.  Motivated by our experience working with real-life databases, we designed \benchName\ to encompass four types of ambiguity. Each entry is designed so that both alternatives have a similar relevance to the question, and a well-calibrated decoding method is expected to rank them close by in their outputs.

We create \benchName\ by modifying the SPIDER \cite{SPIDER} dataset, and use ChatGPT \cite{ChatGPT} to aid with the creation. In each case, we modify the schema instead of the text as that provides greater control over the modification process. We explain the kinds of ambiguity in \benchName\ below and portray examples of each in \tref{tab:BASE}. 

\xhdr{Column Ambiguity (C)} Unlike the SPIDER benchmark where column names usually appear verbatim in the question text (like \texttt{born state} for the column \texttt{born_state}), when users unaware of the schema pose a natural question, they introduce column ambiguity~\cite{ambiguous}. For example, ``\emph{What is the capacity of O2 Arena?}'' could be ambiguous if the schema has separate columns for standing and seating capacity. Likewise, a query on the number of under-nourished children is ambiguous if we have different columns for ``under-weight children'' and ``stunted growth in children''. 

To simulate column ambiguity, for each text $\vx$, schema $\schema$, and SQL $\vy$ in SPIDER, we prompt ChatGPT to generate two synonyms for each column name of $\schema$ in a one-shot manner. Appendix~\ref{app:promptdetails} furnishes more details of the prompt. We then modify $\vs$ by replacing $c$ with two columns $c_1,c_2$, and we use $\vy$ to generate two queries $\vy_1,\vy_2$ where all mentions of $c$ are replaced with $c_1$ in $\vy_1$ and with $c_2$ in $\vy_2$.  An example appears in the first row of \tref{tab:BASE}. We do not reuse $c$ because the SPIDER dataset often contains column names verbatim in the question, and that would violate our attempt at keeping the two options at similar relevance levels. We modify one column at a time and generate up to $3$ examples from each original entry. 
    
\xhdr{Table Ambiguity (T)} Table name ambiguity is common in databases obtained by integrating multiple data sources, as in web tables~\cite{Cafarella08Web,pimplikar12}. Here again, we prompt ChatGPT to generate two alternate names for each table. We then modify one SQL $\vy$ to generate two candidates $\vy_1,\vy_2$ as shown in Table~\ref{tab:BASE}. 

\xhdr{Join Ambiguity (J)} In production databases, a logical table is often vertically partitioned across several tables for efficient clustered access~\cite{StonebrakerABCCFLLMOORTZ14}.  Column names overlapping across tables leads to Join Ambiguity. Suppose we have two tables: (1) \texttt{person} with columns \texttt{id, name, email\_address}, and (2)  \texttt{person\_details} with columns \texttt{id, postal\_address, photo}. A question asking for a person's name and address is ambiguous on whether a \texttt{JOIN} with the \texttt{person\_details} is necessary. We expose such ambiguity by modifying the schema as follows.

Consider a $(\vx,\schema,\vy)$ triplet. Suppose $\vy$ involves selecting two or more columns $c_1, c_2, \ldots$, not necessarily in the same order, from a table $t$. Suppose further that $c_1$ is not a primary key of $t$. We create a table called $t\_c_1$ that includes just the primary key $pk_t$ of $t$, and $c_1$. The first alternative $\vy_1$ is $\vy$ and the second alternative $\vy_2$ uses a join over $t$ and $t\_c_1$, with everything else staying the same as $\vy$.
    
\xhdr{Precomputed Aggregates (P):} 
This ambiguity is particularly common in data warehouses such as \citet{DataCommons} which pre-aggregate certain variables. For instance, the ``\emph{total rice production}'' of a state might refer to the column \texttt{rice\_production} of \texttt{state} rather than a sum over it. \texttosql\ models have a bias toward introducing a \texttt{sum()...group-by} clause every time \texttt{total} appears in the text. The non-aggregated alternative is usually missing in the top-$k$ options. We incorporate this ambiguity as follows.

For each  $(\vx,\schema,\vy)$, where $\vy$ has at least one aggregate, we construct a new table $t'$. For each aggregate $\mathcal{A}$ over column $c$ in $\vy$, we add to $t'$ the columns $\mathcal{A}'\_c$ for all $\mathcal{A}' \in \{\texttt{avg},\texttt{sum},\texttt{min},\texttt{max}\}$, and the columns grouped by in $\vy$. For \texttt{count(*)} we add a column called \texttt{number}.  We get two gold queries, the original $\vy$   and a second with the group-by replaced by a direct \texttt{SELECT} on $t'$ as shown in the example in Table~\ref{tab:BASE}.  We also support aggregates across multiple tables but skip the details here.

%If the tables in $\vy$ are $t_1,t_2,\ldots,t_k$ and columns aggregated over are $c_1,c_2,\ldots,c_l$, we name $t'$ as $t_1\_t_2\_\ldots\_t_k\_c_1\_\ldots\_c_l$. For example, given \texttt{SELECT t1.name, SUM(t2.attendance) FROM singer as t1 JOIN concert AS t2 ON t1.singer\_id = t2.singer\_id GROUP BY t1.name}, the new table is called \texttt{singer\_concert\_attendance} and contains \texttt{name, avg\_attendance, sum\_attendance, min\_attendance} and \texttt{max\_attendance}. There are two gold queries - $\vy$, and $\vy$ with the aggregate and group-by replaced by a direct \texttt{SELECT} on $t'$.

\section{Are Existing Text-to-SQL systems resilient to ambiguity?}

We evaluate several SOTA \texttosql\ models and decoding algorithms on their ability to generate the alternatives of \benchName\ in their top-$k$ outputs. Descriptions of the systems compared and evaluation metrics appear in Subsection~\ref{sub:methods}. \tref{tab:benchresults2} features the results we obtained.

For all systems, the top-$5$ outputs contain both outputs only for a small percentage of the instances. To investigate the reasons for their poor coverage, we manually inspected several outputs of T5-3B and ChatGPT. A few anecdotes for each kind of ambiguity are shown in Appendix~\ref{ap:examples}.  The reason for the failure is that Beam Search tends to produce outputs that are minor tweaks of the best hypothesis, as also corroborated by prior work~\cite{BSNotGood1, BSNotGood2, BSNotGood3, BSNotGood4}. One example from the `C' split of \benchName\ that illustrates this is displayed in \fref{fig:bserrors}. Recent diversity-promoting decoding strategies like Nucleus~\cite{NucleusSampling} and Typical \cite{TypicalSampling} sampling are designed for natural language and are ineffective for capturing the structural diversity that SQL variants require. These observations motivated the design of our inference algorithm, \algName.

\section{Our method: \algName}
\algName\ attempts to induce  \emph{meaningful} diversity, while steering clear of vacuous forms of diversity in the formatting of the SQL. We first attempt to understand the type of logical diversity required by analyzing the errors of the top-$1$ output of T5-3B on the SPIDER benchmark.
% Changed SOTA -> T5-3B; do we want to say SOTA?

The mistakes of the top-$1$ output are cataloged in \tref{tab:errors}. Apart from the column selection order, which is arguably not a serious error, the top four errors are a wrong number of joins, columns, and incorrect column and table names.  A large fraction of the errors involves the ``skeletal structure'' of the SQL, whereas vanilla Beam Search exhibits little diversity in the SQL structure.  Most of its diversity is around generating alternate forms of string literals, tweaking comparison orders, or swapping the names of temporary variables (like \texttt{t1} with \texttt{t2}).  

\begin{table}[H]
    \begin{small}
    \centering
    \begin{tabular}{l|r}
    \toprule
    Error Type & \makecell{Contrib-\\ution (\%)}\\
    \midrule
    Correct Output & 70.31\\
    Wrong selection order & 7.44\\
    Missing \texttt{JOIN}(s) & 6.09\\
    Missing column(s) & 3.48\\
    Extra \texttt{JOIN}(s) introduced & 2.80\\
    Incorrect column or table names & 2.51\\
    WHERE clause missing or incorrect & 1.35\\
    Extra column introduced & 1.06\\
    ORDER BY clause missing or incorrect & 0.48\\
    GROUP BY clause missing or incorrect & 0.39\\
    Wrong comparison & 0.29\\
    DISTINCT missed or introduced & 0.10\\
    Other & 3.68\\
    \bottomrule
    \end{tabular}
    \caption{A catalog of errors on the SPIDER \texttt{dev} split, based on Exact Match (\texttt{EM}), corresponding to the top-$1$ output from a Beam Search with a beam width of $25$. Most errors stem from an incorrect number of \texttt{JOIN}s or \texttt{SELECT}ions, with incorrect schema names being a concern as well.}
    \label{tab:errors}
    \end{small}
\end{table}

%% Algorithm RF

\SetKwComment{Comment}{/* }{ */}
\SetKw{Continue}{continue}
\SetKwFor{For}{for (}{) $\lbrace$}{$\rbrace$}
\RestyleAlgo{ruled}
\begin{algorithm*}[t]
\begin{small}
\caption{Pseudocode for one Beam Extension step of the Restricted Fill-In Algorithm}
\KwData{Beam width $k$, current hypotheses and scores $(\vy_1,s_1),(\vy_2,s_2),\cdots,(\vy_k,s_k)$, template $\vt$, set of all column names $C$ and table names $T$}
\KwResult{The next set of hypotheses with scores $(\vy'_1,s'_1),\cdots, (\vy'_k, s'_k)$}
$H \gets \emptyset$\;
$U \gets \hspace{0.5em}$\texttt{getPossibleFirstToks}$(C) \hspace{0.5em}\cup \hspace{0.5em}$\texttt{getPossibleFirstToks}$(T)$\;
\For{$i = 1;\ i \leq k;\ i = i + 1$}{
    \If{$\neg$\texttt{HypConformsToTemplate}$(\vy_i, \vt)$}{
        \Comment{If this hypothesis violates the template, don't extend it.}
        \Continue;
    }
    \texttt{ncls}$\hspace{0.5em} \gets \hspace{0.5em}$\texttt{getNextTokClass}$(\vy_i, \vt)$\;
    \Comment{Check if we expect to start a column/table name next.}
    \eIf{\texttt{ncls} $\in \hspace{0.5em}$\texttt{\{column, table\}}}{
        \Comment{Allow branching, but restrict options to whitelist.}
        $H_i \gets \hspace{0.5em}$\texttt{getExtensionsWithScores}$(\vy_i, s_i, U)$\; 
    }{
        \Comment{Disallow branching by only choosing the top-scoring extension}
        $H_i \gets \{$\texttt{getTopExtensionWithScore}$(\vy_i, s_i)\}$\;
    }
    $H \gets H \cup H_i$
}
\Return \texttt{getTopKHypothesesWithScore}$(H)$\;
\label{alg:rfalgo}
\end{small}
\end{algorithm*}

These observations drove us to design a two-stage approach.  In the first stage, we generate diverse SQL skeletons (templates) to capture structural diversity, and in the second we fill in the template with schema-diverse alternatives. We illustrate our approach in \fref{fig:entirepipeline}.

\subsection{Plan-based Template Generation}
%% Pipeline image start
\begin{figure*}[t]
    \centering
    \includegraphics[width=\textwidth]{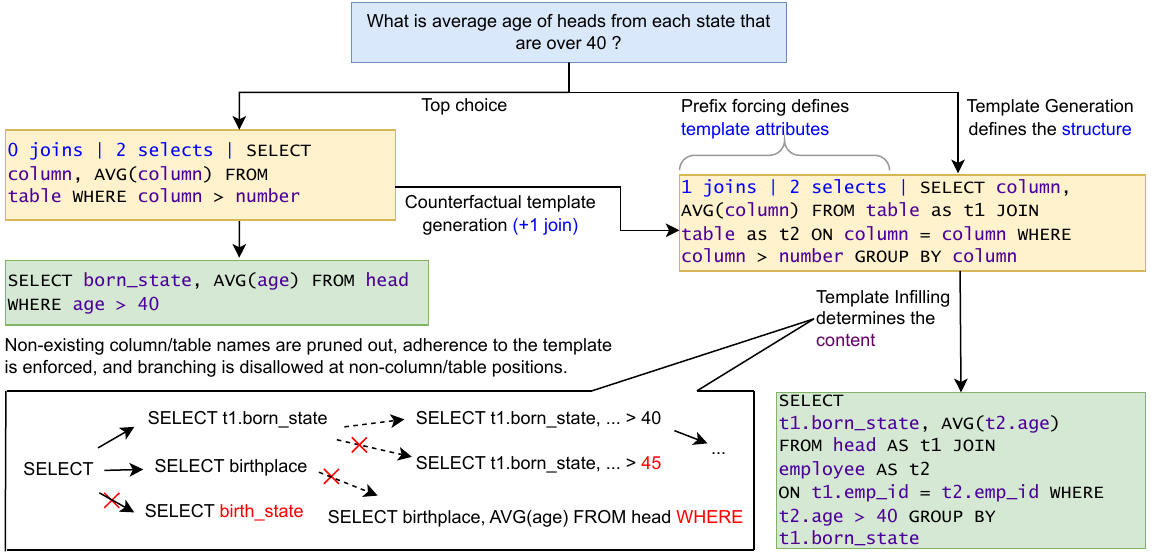}
    \caption{Our approach in its entirety. A counterfactual template generation step provides template diversity via Prefix Enforcement. Constrained infilling generates content diversity by restricting branching and enforcing template adherence.}
    \label{fig:entirepipeline}
\end{figure*}
%% Pipeline image end

%
A template of an SQL query abstracts away the names of the tables and columns of the SQL query, string literals, and constants, so that only the structural components (\texttt{SELECT}s, \texttt{GROUP BY}s, \texttt{JOIN}s, comparisons and so on) remain.    %The concept of templates is explained further in Appendix~\ref{ap:templateinfo}.
On the \texttt{train} split of SPIDER, we convert the gold SQL to a template by a simple rule-based replacement of schema names (details in  Appendix~\ref{ap:templateinfo}) and use it to train a Text-to-Template model. However, the top-$k$ templates found via beam search on this model again lacked logical diversity.
%After training a Text-to-Template model, we found that Beam Search again indulges in meaningless differences and fails to provide enough logical diversity.  
One example is shown by \fref{fig:templateBSdiversity} in Appendix~\ref{ap:decodeinadequate}. We thus explored a more deliberate mechanism to induce diversity following these three steps:

First, we preface a template with a plan declaring the structural properties of the SQL where diversity is desired. 
Based on our error analysis in \tref{tab:errors}, we chose to induce diversity on the number of \texttt{JOIN}s and final \texttt{SELECT}ions. Thus, for a given input question, we output a plan followed by a template as:% We trained the template generation model to predict these numbers as a prefix, which we also refer to as a \emph{plan}:
\begin{center}
    \vspace*{-0.5em}
    \texttt{<J> joins | <S> selects | <TEMPLATE>}
    \vspace*{-0.5em}
\end{center}
The left yellow box in Figure~\ref{fig:entirepipeline} shows one such plan prefixed template.  

Second, we counterfactually perturb the counts in the plan as follows. We generate the top-choice template $t$ without any constraints (say, with $j$ joins and $s$ selections). We then generate four diverse plans by searching in the neighborhood of the most likely predicted structure as $(j-1,s), (j+1,s), (j,s-1), (j,s+1)$.   We skip invalid combinations ($j < 0$, $j > 3$, or $s \leq 0$).  We also explored sampling $j,s$ based on predicted probabilities, but these were extremely skewed. 

Finally, for each plan (enforced as a prefix), we use greedy decoding to generate the template. The decoding algorithm was good at generating templates as per the specified plan.  

Thus, at the end of the template generation phase, we have at most five templates.

%respectively, by enforcing the corresponding plan during Beam Search. Due to the output format, specifying a plan via prefix forcing effectively controls the number of \texttt{JOIN}s and selections in the SQL query. The top-$1$ outputs from each $(j,s)$ combination are passed on to the infilling step. Template generation uses greedy decoding (Beam Width = $1$) and is lightweight. 

\subsection{Template filling with Diverse Schema}
\label{sub:vdpf}
Here we fill diverse column names and table names in the generated templates.
% Having extracted diversity in other ways, we now consider alternate columns and tables to fill in the template. Beam Search is still very cumbersome, as it is stubborn about introducing false diversity by, \emph{e.g.}, omitting spaces, continually repeating a phrase, replacing commas with `\emph{and}', and so on. 
We use beam search to this end but enforce adherence to the template. We track our position in the template during infilling. If the next token is expected to not be part of a table or column name, we disallow the model from exploring anything apart from the highest-scoring next token. Otherwise, we allow it to branch in the next decoding step. However, we restrict its options to a whitelist of tokens computed beforehand by enumerating the columns/tables from the schema. The pseudocode of our Restricted Infilling method is presented in Algorithm~\ref{alg:rfalgo}.

The next challenge is how to rank the SQLs from the diverse templates and select the top-5. We initially attempted to rank based on the product of probabilities of the template and in-filling steps. However, the probability distribution of the models we worked with was extremely skewed - for example, top-$p$ sampling with $p=0.9$ produced the same template in all infillings over $70\%$ of the time. Combined with the well-known lack of calibration of neural sequence models, we found it better to simply choose the top$-2$ SQLs from each template, along with the top$-2$ from vanilla beam-search without any templates.  
After filtering out duplicates, the top-$5$ queries in the list are returned.
%Each template gives us multiple SQL queries after infilling, creating a 2-dimensional grid of outputs. The $(i,j)$-th entry corresponds to the $j$-th infilling of the $i$-th template. We initially experimented with a score-based ranking, where the score was obtained by summing the log probabilities of the template and filled-in query. However, the probabilities of the templates are usually skewed, resulting in one template monopolizing the top ranks. We, therefore, settled instead for the zig-zag enumeration instead (i.e., $(1,1),(1,2),(2,1),\ldots$), as it affords both template and content diversity. We use an approximate but fast way to do this by assigning the $(i,j)$-th output weight of $3i+2j$ and ranking the outputs by increasing weight. 

%The outputs are then combined with the top-$2$ outputs of T5-3B as follows. We rank the outputs as described above and filter out duplicates. Finally, we select the $3$ outputs that rank best to combine with T5-3B's top-$2$ outputs, for a total of $5$ queries.

\section{Experiments}
We present extensive comparisons of several State-of-the-Art \texttosql\ models and decoding methods on \benchName\ in the following sections. We then show that \algName\ can be helpful even in the absence of ambiguity. We also present a detailed ablation study of \algName, and a discussion of its use-cases and shortcomings.   
\begin{table*}[t]
    \begin{small}
    \centering
    \begin{tabular}{c|c|c|c|c|c|c|c|c|c}
    \toprule
    \makecell{Kind of\\Ambiguity} & CGPT & Codex & RSQL & F-T5-3B & T5-3B & T5-3B-k & T5-3B-p & T5-3B-T & \algName\\
    \midrule
    \multicolumn{10}{c}{\Any\ (\%)}\\
    \midrule
    C & 52.7 & 55.6 & 51.0 & 60.7 & 59.0 & 52.0 & 51.7 & 48.1 & \textbf{66.6}\\
    T & 55.7 & 59.8 & 33.8 & 59.7 & 57.9 & 49.1 & 48.7 & 43.6 & \textbf{67.3}\\
    J & 77.8 & 83.7 & 68.8 & 86.1 & 86.8 & 80.6 & 79.9 & 76.7 & \textbf{87.2}\\
    P & 57.4 & \textbf{77.2} & 42.6 & 49.5 & 58.4 & 55.4 & 53.5 & 51.5 & 64.4\\
    \midrule
    \multicolumn{10}{c}{\All\ (Coverage) (\%)}\\
    \midrule
    C & 22.7 & 10.4 & 10.8 & 8.7 & 11.7 & 3.3 & 2.5 & 0.0 & \textbf{28.0}\\
    T & 37.3 & 14.3 & 6.4 & 15.1 & 21.9 & 8.0 & 7.1 & 1.1 & \textbf{42.6}\\
    J & 15.6 & 43.8 & 0.0 & 22.6 & 27.8 & 2.8 & 1.7 & 0.0 & \textbf{59.4}\\
    P & 8.9 & 24.7 & 8.9 & 2.9 & 15.8 & 4.0 & 3.0 & 0.0 & \textbf{24.8}\\
    \bottomrule
    \end{tabular}
    \caption{The results of all compared systems on \benchName\, as portrayed by Execution Match (\texttt{EXM}) accuracy in the Top-5 outputs. \algName\ usually performs the best under the \Any\ heading, except for Precomputed Aggregates. More importantly, \algName\ consistently outperforms all other systems under the \All\ heading.  This shows the capacity of \algName\ to capture greater meaningful diversity in its outputs.}
    \label{tab:benchresults2}
    \end{small}
\end{table*}
\subsection{Implementation Details of \algName}
Both stages of LogicalBeam are fine-tuned versions of Flan T5-3B (\texttt{max length} $= 512$), with an Adafactor \cite{Adafactor} optimizer (learning rate $1e-4$, and no decay). The models were trained for roughly $300$ epochs each, with checkpoint selection based on the highest Template Match and Exact match, respectively (on the validation set, with greedy decoding). Our datasets for the models consist of one-to-one maps of each example from SPIDER, with, e.g., the SQL query replaced by the corresponding template for the Text-to-Template model. We use the HuggingFace \texttt{LogitsProcessor}\footnote{\url{https://huggingface.co/docs/transformers/internal/generation_utils\#logitsprocessor}} for the Template-Infilling model, which allows us to modify logits at each decoding step. We set all the disallowed tokens’ logits to $-\infty$ to implement the restricted beam search.
\subsection{Methods Compared}
\label{sub:methods}
We compare with the following models. All use Beam Search with a beam width of $10$ unless otherwise specified. %\algName\ also uses a beam width of $10$ during infilling. 
For T5-3B (one of the best-performing baselines), alternate decoding algorithms are also included in the comparison. 

\xhdr{ChatGPT (CGPT):} We prompt ChatGPT for its top five choices given the question and schema in a one-shot manner using an example outside of \benchName. One-shot prompting was required to get ChatGPT to adhere to the output format. More details can be found in Appendix~\ref{app:promptdetails}. We also show in Appendix~\ref{ap:altprompt} that alternate prompts tried by prior works (such as \cite{ChatGPTEval}) are inefficient in getting ChatGPT to cover all possibilities. 

\xhdr{OpenAI Codex (Codex):} We use few-shot prompting with the \texttt{code-davinci-002} version of OpenAI Codex \cite{codex}. This is the most capable Codex version at the time of writing. More details are provided in Appendix~\ref{app:promptdetails}. 

\xhdr{RESDSQL (RSQL):} Among approaches that do not use ChatGPT/GPT-4, RESDSQL~\cite{resdsql} is the best-performing method on SPIDER at the time of writing. We use its 3B variant (the most potent one) for comparison but turn off the NatSQL \cite{NatSQL} representation, as it is orthogonal to our approach and can be used with it. 

\xhdr{T5-3B (T5-3B):} We use the T5-3B checkpoint from the PICARD \cite{PICARD} repository that fine-tunes T5-3B on SPIDER. By default, we use Beam Search for T5-3B.

\xhdr{T5-3B with Top-$k$ sampling (T5-3B-k):} At each step of decoding, we sample from the top-$50$ tokens, i.e. using top-$k$ sampling with $k=50$. 

\xhdr{T5-3B with Nucleus/Top-$p$ Sampling (T5-3B-p):} At each step of decoding, we sample from the top-$p$ tokens that account for 90\% of the probability mass as proposed in~\cite{NucleusSampling}.

\xhdr{T5-3B with Typical Sampling (T5-3B-T):} Typical Sampling~\cite{TypicalSampling} is another recent diverse decoding algorithm for enforcing natural diversity. This algorithm uses a parameter, \texttt{typical\_p}, similar to the \texttt{top\_p} of Nucleus Sampling. Following \cite{TypicalSampling}, we set \texttt{typical\_p} to $0.9$.

\xhdr{Flan T5-XL (F-T5-3B):} This is the FLAN variant of the T5-3B model, fine-tuned with the same PICARD code as the T5-3B model above.

\xhdr{\algName} For both stages we fine-tuned 
separate Flan T5-3B \cite{Flan-T5} models.  We use a learning rate of $1\cdot10^{-4}$ and an effective batch size of $810$ via gradient accumulation in both cases. 

\xhdr{Evaluation Metrics}
We present two types of accuracies (i) \emph{\Any} - that checks if either of the gold queries feature in the top-$5$ outputs (ii) \emph{\All} - that checks if both gold queries feature in the top-$5$.  We only report the Execution Match (\texttt{EXM}) accuracies for each. The numbers of Exact Set Match are given in Appendix~\ref{ap:fullresults}.

\subsection{Overall comparison on \benchName}
\label{sub:comp}

%% Scaling start

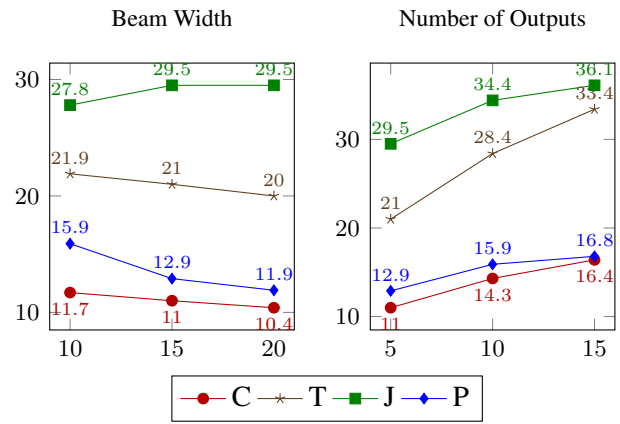
\begin{figure}[t]
\centering
\hspace*{-0.8em} % Fix margins for final version
\begin{tikzpicture}
\begin{groupplot}[group style={group size= 2 by 1,ylabels at=edge left},width=0.3\textwidth,height=0.32\textwidth]
\nextgroupplot[label style={font=\small},
tick label style={font=\small},
legend style={at={($(0,0) + (1cm, 1cm)$)},legend columns=6,fill=none,draw=black,anchor=center,align=center},
legend to name=fred,
title = \makecell{\small Beam Width},
mark size=2pt]
\addplot[red!70!black,mark=*,nodes near coords align={below}, nodes near coords style={font=\scriptsize}] coordinates {(10, 11.7) (15, 11.0) (20, 10.4) };
\addplot[brown!50!black,mark=star, nodes near coords style={font=\scriptsize}] coordinates {(10, 21.9) (15, 21.0) (20, 20.0) };
\addplot[green!50!black,mark=square*, nodes near coords style={font=\scriptsize}] coordinates {(10, 27.8) (15, 29.5) (20, 29.5) };
\addplot[blue,mark=diamond*, nodes near coords style={font=\scriptsize}] coordinates {(10, 15.9) (15, 12.9) (20, 11.9) };
\addlegendentry{C};    
\addlegendentry{T};    
\addlegendentry{J};          
\addlegendentry{P};    

\nextgroupplot[label style={font=\small},
tick label style={font=\small},
title = \makecell{\small Number of Outputs},
mark size=2pt]
5,11.0,21.0,29.5,12.9 %,8.20
10,14.3,28.4,34.4,15.9 %,9.40
15,16.4,33.4,36.1,16.8 %,10.80
\addplot [red!70!black,mark=*,nodes near coords align={below}, nodes near coords style={font=\scriptsize}] coordinates {(5, 11.0) (10, 14.3) (15, 16.4) };
\addplot [brown!50!black,mark=star, nodes near coords style={font=\scriptsize}] coordinates {(5, 21.0) (10, 28.4) (15, 33.4) };
\addplot [green!50!black,mark=square*, nodes near coords style={font=\scriptsize}] coordinates {(5, 29.5) (10, 34.4) (15, 36.1) };
\addplot [blue,mark=diamond*, nodes near coords style={font=\scriptsize}] coordinates {(5, 12.9) (10, 15.9) (15, 16.8) };
\end{groupplot}
\coordinate (c1) at (rel axis cs:1,1);
\coordinate (c2) at ($(c1) + (0.7cm, 0)$);
\node[below] at (c2 |- current bounding box.south)
{\pgfplotslegendfromname{fred}};

\end{tikzpicture}
\caption{The coverage only increases slightly with more outputs, and \emph{decreases} with increasing beam width. The x-axis varies the controlled hyperparameter, while the y-axis reports coverage.}
\label{fig:bwandno}
\end{figure}

%% Scaling end

%% Table ablation start
\begin{table}[t]
    \begin{small}
    \centering
    \setlength{\tabcolsep}{4pt}
    \begin{tabular}{c|c|c|c|c}
    \toprule
    \makecell{Kind of\\Ambig-\\uity} & \makecell{Single\\stage} & \makecell{Two\\stages} & \makecell{+Template\\Diversity} & \makecell{+Schema\\Diversity\\(\algName)} \\
    \midrule
    \multicolumn{5}{c}{\Any\ (\%)}\\
    \midrule
    C & 64.0 & 65.9 & 65.9 & \textbf{66.6}\\
    T & 60.7 & 66.2 & 65.0 & \textbf{67.3}\\
    J & 86.8 & \textbf{88.5} & 87.1 & 87.2\\
    P & 58.4 & 62.4 & 63.4 & \textbf{64.4}\\
    \midrule
    \multicolumn{5}{c}{\All\ (Coverage) (\%)}\\
    \midrule
    C & 23.2 & 16.0 & 16.1 & \textbf{28.0}\\
    T & 25.3 & 28.4 & 28.2 & \textbf{42.6}\\
    J & 54.5 & 54.5 & \textbf{62.2} & 59.4\\
    P & 9.9 & 27.7 & \textbf{30.7} & 24.8\\
    \bottomrule
    \end{tabular}
    \caption{The Execution Match (\texttt{EXM}) accuracies of the ablations on \benchName. Template Diversity helps with Join Ambiguity and Precomputed Aggregates, while Schema Diversity aids with Column/Table ambiguity.}
    \label{tab:benchresults3}
    \end{small}
\end{table}

We present the results of the system comparison in \tref{tab:benchresults2}.  We make the following observations:
\begin{itemize}[leftmargin=*]
    \item \textbf{State-of-the-art Text-to-SQL models cannot handle ambiguity:} Existing approaches, including T5-3B, ChatGPT, and RESDSQL among others, fail to cover both alternatives in Top-5 even when they perform reasonably under the \emph{\Any} heading. Surprisingly, despite being SOTA on the SPIDER dataset, RESDSQL sees its coverage plummet under ambiguity. We observed that it often produced outputs that corresponded to neither of the alternatives. This behavior was also exhibited by T5-3B, by using aggregates such as \texttt{max(avg\_age)}. Though outputs produced this way are syntactically correct, they do not correspond to any meaningful question.
  \item \textbf{Beam-search gives unhelpful token-level diversity} %Beam Search is inadequate to draw out sufficient meaningful diversity and  
  Although it may seem like increasing the beam width allows greater exploration and thus greater diversity, this is not the case. As \fref{fig:bwandno} shows, increasing beam width generally \emph{reduces} coverage since the model is naturally biased towards one of the two alternatives, and a greater beam width only serves to let the model discover more vacuous variants of it. Using a larger number of outputs does not help much, as attested to by \fref{fig:bwandno}. Even a $3\times$ increase in the number of outputs leads only to marginal improvements, except for Table Ambiguity. 
    \item \textbf{Recent diversity-promoting decoding algorithms fail due to skewed token distribution}  SOTA diversity-promoting alternatives to Beam Search such as Nucleus and Typical sampling perform \emph{worse} than beam search. Although these have demonstrated strong performance for tasks such as text summarization~\cite{TypicalSampling}, they cannot promote meaningful diversity in semantic parsing under ambiguity since the model produces skewed token probabilities. This causes the sampled hypotheses to be often identical as seen in the anecdotes in \fref{fig:nserrors} in the supplementary material.  
\item \textbf{\algName\ yields substantial accuracy in covering both ambiguous outputs}. Its \All\ accuracy is almost \textbf{2.5$\times$} better in the case of Column Ambiguity than T5-3B, and \textbf{2$\times$} in the case of Table ambiguity. It outperforms other systems by a huge margin. The accuracy under Join Ambiguity increases by \textbf{1.4}$\times$ over Codex (the next-best method) and is over \textbf{2}$\times$ better than any other method. For Precomputed Aggregates, \algName\ is once again the best system, and surpasses everything apart from Codex by over \textbf{1.5}$\times$. We also convincingly beat ChatGPT and OpenAI Codex across the board on coverage.      
\end{itemize}
\subsection{Performance on Unambiguous Queries}
Although our main focus is coverage under ambiguity, we also evaluate our proposal against the baseline T5-3B model on the \texttt{dev} split of SPIDER. We find that \algName\ doesn't just help the \benchName\ benchmark but also provides gains on conventional \texttosql\ benchmarks like SPIDER where ambiguity is limited. \tref{tab:nonamb-results} shows that \algName\ improves the top-$5$ Exact-Set and Execution Match accuracies on SPIDER by \textbf{2.3\%} and \textbf{3.1\%} over the baseline, respectively. As another example, we evaluate our method on the \texttt{dev} split of the challenging Kaggle DBQA \cite{kaggleDBQA} benchmark. We observe a drastic increase in the top-$5$ Exact-Set and Execution Match accuracies, from \textbf{27.1\%} and \textbf{26.5\%} to \textbf{35.4\%} and \textbf{35.4\%}, respectively. We conclude that \algName\ is useful across a wide range of Semantic Parsing tasks. Unlike earlier grammar-based generators like SmBoP~\cite{rubin-berant-2021-smbop-semi} that require special decoder models, our approach can work within existing LM-based models. 

%% Table comparison on SPIDER and KDBQA
\begin{table}[t]
    \begin{small}
    \centering
    \setlength{\tabcolsep}{4pt}
    \begin{tabular}{c|c|c}
    \toprule
    \makecell{Method} & \makecell{Top-5 Exact Set\\Match} & \makecell{Top-5 Execution\\Accuracy}\\
    \midrule
    \multicolumn{3}{c}{SPIDER (\texttt{dev} split, \%)}\\
    \midrule
    T5-3B  & 76.1 & 78.2\\
    LogicalBeam & \textbf{78.4} & \textbf{81.3}\\
    \midrule
    \multicolumn{3}{c}{Kaggle DBQA (\texttt{dev} split, \%)}\\
    \midrule
    T5-3B  & 27.1 & 26.5\\
    LogicalBeam & \textbf{35.4} & \textbf{35.4}\\
    \bottomrule
    \end{tabular}
    \caption{The Exact-Set and Execution Match accuracies of LogicalBeam on two popolar \texttosql datsets, SPIDER and Kaggle DBQA. Despite the datasets not exhibiting ambiguity, LogicalBeam delivers significant improvements over the T5-3B baseline.}
    \label{tab:nonamb-results}
    \end{small}
\end{table}

\subsection{Ablation study}
\label{sub:ablation}   
\algName\ has three design decisions: (1) Use of a two-step approach, (2) Counterfactual structural directives via plans, (3) Template-guided schema diversity.  We present an ablation study where we incrementally add these changes in Table~\ref{tab:benchresults3}. The first column (``Single Stage'') generates an SQL directly with a prefix for structural diversity, differing from \algName\ only in using a single stage. It still uses plan enforcement and branching control. We find that its coverage lags behind \algName, and by a large margin for T and P.  The primary reason could be that template-guided decoding allows us to discard erroneous extensions at \emph{each} decoding step. %This benefit accrues exponentially and steers us clear of unhelpful outputs. For example, if at each step our search space is smaller by $90\%$ on average, the net space shrinks by $>500\times$ over $60$ steps. For $100$ steps, the factor is over $35000$.
The second column (``Two Stages'') shows a simple two-stage method where we generate a template without any counterfactual control, and use Beam Search to fill it in. This method decouples template and schema diversity, but cannot encourage either by itself. Forcing counterfactual diversity (``+Template Diversity'') boosts the coverage under Join Ambiguity and Precomputed Aggregates. Finally, encouraging Schema Diversity via our Restricted Fill-In Algorithm (\algName, the last column) significantly improves coverage for Column and Table Ambiguity.

\subsection{Discussion}
\algName\ is general and need not be confined to the world of Semantic Parsing. For instance, the plan (prefix) could be replaced with any aspect of a code snippet that we wish to control. More generally, since the underlying mechanism only involves the model being faithful to the prefix and has no manual components, we could do the same with almost any Sequence-to-Sequence task (for example, political alignment in news summarization).

\algName\ consistently improves performance both under ambiguity and in the absence of it, often by drastic margins. However, we would also like to highlight one failure mode we observed, that was also exhibited by other approaches. Consider a query ``\texttt{... table1 as t1 JOIN table2 as t2}''. On rare occasions, we observed that an identical query with \texttt{t2} replaced by \texttt{t3} (and \texttt{t2} skipped) was also present in the choices. We believe this indicates a strong bias of the underlying model towards a particular template – so much so that it prefers this weird \texttt{(t1, t3)} combination to introducing template diversity. The problem of debiasing the model makes for exciting future work. It is not unique to Semantic Parsing, and, we believe, deserves attention in its own right.

\section{Conclusion}
In this work, we highlighted the lack of evaluation of \texttosql\ models under ambiguity in contemporary literature. To address this, we developed \benchName, a novel benchmark with 3000+ challenging examples that evaluates \texttosql models on four kinds of ambiguity. We demonstrated that current methods fall short of acceptable performance under ambiguity. Motivated by analyzing the errors of a T5-3B model on the SPIDER dataset, we developed a two-step approach of generating and then filling in a template. To this end, we trained a model to predict the number of \texttt{JOIN}s and selections as a plan before the template, and controlled template diversity by enforcing appropriate plans. Beam Search was modified to enforce template adherence during in-filling. Our method aligns well with intuition and greatly improves a model's coverage under ambiguity, as measured on \benchName. It also delivers improvements in the absence of ambiguity, on the SPIDER and Kaggle DBQA datasets. We hope our efforts inspire future work to study generation under ambiguity in more detail, both in the domain of \texttosql conversion and beyond.  

\section*{Limitations}
In this work, we curated a benchmark of ambiguous queries by perturbing SPIDER, an existing dataset. While we believe that our benchmark is a good measure of performance under ambiguity, real-life databases may exhibit more numerous as well as varied forms of ambiguity. In addition, \benchName\ only consists of examples with questions in English. Ambiguity may manifest differently based on the choice of natural language, and a corresponding study should make for interesting future work.

Due to the two-step approach, \algName\ incurs a higher number of decoding steps as compared to an end-to-end model. However, due to using a lightweight Greedy Search for the first stage, the number of decoding steps of \algName\ falls not much beyond the baseline. Nevertheless, finding an optimal trade-off between decoding steps and coverage remains an intriguing challenge.

At the time of writing, ChatGPT and OpenAI Codex represent the most powerful publicly available LLMs suitable for Text-to-SQL conversion and are unable to exhibit sufficient diversity under ambiguity. Future versions or models may overcome this barrier. 

% EMNLP 2023 requires all submissions to have a section titled ``Limitations'', for discussing the limitations of the paper as a complement to the discussion of strengths in the main text. This section should occur after the conclusion, but before the references. It will not count towards the page limit.  

% The discussion of limitations is mandatory. Papers without a limitation section will be desk-rejected without review.
% ARR-reviewed papers that did not include ``Limitations'' section in their prior submission, should submit a PDF with such a section together with their EMNLP 2023 submission.

% While we are open to different types of limitations, just mentioning that a set of results have been shown for English only probably does not reflect what we expect. 
% Mentioning that the method works mostly for languages with limited morphology, like English, is a much better alternative.
% In addition, limitations such as low scalability to long text, the requirement of large GPU resources, or other things that inspire crucial further investigation are welcome.

\section*{Ethics Statement}
As we generate \benchName\ by perturbing SPIDER \cite{SPIDER}, it does not contain any identifying information. We do not foresee any broader ethical impacts of our work.

% Scientific work published at EMNLP 2023 must comply with the \href{https://www.aclweb.org/portal/content/acl-code-ethics}{ACL Ethics Policy}. We encourage all authors to include an explicit ethics statement on the broader impact of the work, or other ethical considerations after the conclusion but before the references. The ethics statement will not count toward the page limit (8 pages for long, 4 pages for short papers).

\section*{Acknowledgements}
We thank Microsoft for sponsoring access to Azure OpenAI via the Accelerate Foundation Models Academic Research Initiative. We also extend our heartfelt gratitude to the Anonymous Reviewers who helped improve the paper with their insightful observations and suggestions. 

\bibliography{anthology, references}
\bibliographystyle{acl_natbib}

\appendix
\begin{lstlisting}[float=*, caption={The directive we use for asking ChatGPT to produce table synonyms.}, label={lst:coldirective}, captionpos=b, belowskip=-0.8 \baselineskip]
You are a helpful assistant that assists the user in deciding alternate names for their tables in an SQL database.
\end{lstlisting}
\begin{lstlisting}[float=*, caption={The prompt we use for asking ChatGPT to produce table synonyms.}, label={lst:colprompt}, captionpos=b, belowskip=-0.8 \baselineskip]
The database with database ID "[DB_ID]" currently has tables with the names [TABLES_STRING]. Give me two alternate names for the table "[TABLE_NAME]". Print your output as a python list. Do not print any additional information, formatting, explanation, or notes.
\end{lstlisting}
\begin{lstlisting}[float=*, caption={The directive we use for asking ChatGPT to produce column synonyms.}, label={lst:tbldirective}, captionpos=b, belowskip=-0.8 \baselineskip]
You are a helpful assistant that assists the user in deciding alternate names for their tables' columns in an SQL database.
\end{lstlisting}
\begin{lstlisting}[float=*, caption={The prompt we use for asking ChatGPT to produce column synonyms.}, label={lst:tblprompt}, captionpos=b, belowskip=-0.8 \baselineskip]
The database with database ID "[DB_ID]" has a table called "[TABLE_NAME]". This table has columns with the following names:
[COLUMN_NAMES]
Give me two alternate names for each column. Format your output as a json snippet with keys corresponding to column names. Do not print any additional information, formatting, explanation, or notes.
\end{lstlisting}
\section{Computational Resources and Prompting Details}
\label{app:promptdetails}
We highlight in this section the prompts we used for prompting ChatGPT, both for the synonyms of table/column names and for the Text-to-SQL conversion on \benchName. We also provide the prompts we used with OpenAI Codex, and furnish details of the computational resources we used. All details provided below are specified as of June 20, 2023.
\subsection{Computational Resources}
All of our experiments were run on a single NVIDIA A100 GPU with 80GB of memory. We estimate the total GPU usage to have been roughly 500 GPU hours across training and inference. We further estimate the cost of utilizing ChatGPT and OpenAI Codex to be under $100\$$ in total.
\subsection{Synonyms (ChatGPT)}
For column and table synonyms, we use one-shot prompting to indicate to ChatGPT the kind of transformation we desire. 

For column synonyms, the overall directive and prompt are shown in Listings~\ref{lst:coldirective} and \ref{lst:colprompt} respectively. The demonstrated example also follows the format of the prompt.
\texttt{[DB\_ID]} is the database ID of the database having the column, and \texttt{[TABLE\_NAME]} is the name of the table containing it. A comma-separated list of all database table names in quotes is filled into \texttt{[TABLES\_STRING]}.

Similarly, for table synonyms, the directive and prompt are shown in Listings~\ref{lst:tbldirective} and \ref{lst:tblprompt} respectively. In particular, \texttt{[DB\_ID]} and \texttt{[TABLE\_NAME]} are replaced with the database ID and table name. \texttt{[COLUMN\_NAMES]} is a comma-separated list of columns of the specified table. The demonstrated example also follows this format.

We found that asking ChatGPT to structure its output as a JSON snippet saved us the trouble of sanitizing its outputs and separating it from any decoration (comments or explanation) it produced. It also made it easier to detect invalid outputs and retry.
\subsection{Text-to-SQL (ChatGPT)}
We prompted ChatGPT in a one-shot manner for evaluation on our benchmark. This was necessary as our benchmark is built by modifying SPIDER \cite{SPIDER}. The queries are expected to be in a specific format in the spider dataset. In particular, the table aliases are always \texttt{t1, t2, \ldots}. Further, columns are never aliased, and only unqualified \texttt{JOIN} is used and \texttt{INNER JOIN}s, \texttt{OUTER JOIN}s not used. Therefore, rather than do some ad-hoc post-correction, we showed ChatGPT one example from the original SPIDER \texttt{dev} set.
In addition, we asked ChatGPT to structure its output as a JSON snippet, a departure from the conventional prompt as in \cite{ChatGPTEval}. This was motivated by our observation that ChatGPT would occasionally sneak comments or notes into its queries despite our best efforts. By asking it to produce the output in a structured (JSON) format, it was much easier to detect errors and retry.

\begin{lstlisting}[float=*, caption={The directive we use while prompting ChatGPT on our benchmark.}, label={lst:benchdirective}, captionpos=b, belowskip=-0.8 \baselineskip]
You are a helpful assistant that converts provided English questions to SQL queries with respect to a provided schema.
\end{lstlisting}
\begin{lstlisting}[float=*, caption={The prompt we use while prompting ChatGPT on our benchmark.}, label={lst:benchprompt}, captionpos=b, belowskip=-0.8 \baselineskip]
The schema for a database with Database ID [DB_ID] is:
[SCHEMA]
Convert the following English question to the five most plausible SQL queries compatible with the above schema.
Use simply the column name for selections in simple queries. For queries with joins, use t1, t2, and so on as aliases for the tables, and use t1.column, t2.column, and so on for the column selections.
Structure your output as a JSON snippet with a single key "queries", mapping to a list of alternatives. Do not print any additional information, explanation, formatting, or notes.
Question: [QUESTION]
\end{lstlisting}

%% Table EM
\begin{table*}[t]
    \begin{small}
    \centering
    \begin{tabular}{c|c|c|c|c|c|c|c|c|c}
    \toprule
    \makecell{Kind of\\Ambiguity} & CGPT & Codex & RSQL & F-T5-3B & T5-3B & T5-3B-k & T5-3B-p & T5-3B-T & \algName\\
    \midrule
    \multicolumn{10}{c}{\Any\ (\%)}\\
    \midrule
    C & 43.4 & 51.5 & 49.4 & 58.6 & 57.6 & 52.3 & 51.9 & 48.2 & \textbf{65.8}\\
    T & 41.1 & 54.3 & 30.2 & 58.9 & 56.5 & 47.6 & 47.3 & 41.9 & \textbf{66.8}\\
    J & 67.4 & 82.6 & 68.1 & 87.5 & 85.4 & 82.3 & 82.3 & 79.5 & \textbf{89.2}\\
    P & 51.5 & 80.2 & 57.4 & 65.4 & 78.2 & 72.3 & 74.3 & 69.3 & \textbf{81.2}\\
    \midrule
    \multicolumn{10}{c}{\All\ (Coverage) (\%)}\\
    \midrule
    C & 19.8 & 9.4 & 10.5 & 8.8 & 12.2 & 3.2 & 2.6 & 0.0 & \textbf{27.7}\\
    T & 28.9 & 11.6 & 4.7 & 14.3 & 20.6 & 6.5 & 5.7 & 0.0 & \textbf{39.9}\\
    J & 15.3 & 41.0 & 0.0 & 22.6 & 24.3 & 2.8 & 1.7 & 0.0 & \textbf{57.3}\\
    P & 7.9 & \textbf{23.8} & 7.9 & 3.0 & 16.8 & 4.0 & 2.0 & 0.0 & 22.8\\
    \bottomrule
    \end{tabular}
    \caption{The Exact Set Match (\texttt{EM}) Accuracy of the compared systems.}
    \label{tab:systemsEM}
    \end{small}
\end{table*}

% %% Table ablation EM [OLD]
% \begin{table}[t]
%     \begin{small}
%     \centering
%     \setlength{\tabcolsep}{4pt}
%     \begin{tabular}{c|c|c|c|c|c}
%     \toprule
%     \makecell{Kind of\\Ambig-\\uity} & Direct & BS+NS & BS+BS & PE+BS & PE+RF\\
%     \midrule
%     \multicolumn{5}{c}{\Any\ (\%)}\\
%     \midrule
%     C & 64.3 & 62.8 & 64.2 & 63.6 & \textbf{64.8}\\
%     T & 60.0 & \textbf{64.6} & \textbf{64.6} & 63.2 & 63.4\\
%     J & 88.9 & \textbf{90.3} & \textbf{90.3} & 89.2 & 89.6\\
%     P & 78.2 & 79.2 & \textbf{80.2} & 78.2 & \textbf{80.2}\\
%     \midrule
%     \multicolumn{5}{c}{\All\ (Coverage) (\%)}\\
%     \midrule
%     C & 23.5 & 12.2 & 16.0 & 16.0 & \textbf{29.4}\\
%     T & 20.7 & 21.1 & 26.8 & 26.6 & \textbf{40.2}\\
%     J & 53.1 & 56.2 & 56.6 & \textbf{64.6} & 59.0\\
%     P & 7.9 & 28.7 & 27.7 & \textbf{29.7} & 24.7\\
%     \bottomrule
%     \end{tabular}
%     \caption{The Exact Set Match (\texttt{EM}) accuracies of the compared ablations on \benchName.}
%     \label{tab:ablationEM}
%     \end{small}
% \end{table}
% %% EM

%% Table ablation EM
\begin{table}[t]
    \begin{small}
    \centering
    \setlength{\tabcolsep}{4pt}
    \begin{tabular}{c|c|c|c|c}
    \toprule
    \makecell{Kind of\\Ambig-\\uity} & \makecell{Single\\stage} & \makecell{Two\\stages} & \makecell{+Template\\Diversity} & \makecell{+Schema\\Diversity\\(\algName)} \\
    \midrule
    \multicolumn{5}{c}{\Any\ (\%)}\\
    \midrule
    C & 64.3 & 64.2 & 63.6 & \textbf{65.8}\\
    T & 60.0 & 64.6 & 63.2 & \textbf{66.8}\\
    J & 88.9 & \textbf{90.3} & 89.2 & 89.2\\
    P & 78.2 & 80.2 & 78.2 & \textbf{81.2}\\
    \midrule
    \multicolumn{5}{c}{\All\ (Coverage) (\%)}\\
    \midrule
    C & 23.5 & 16.0 & 16.0 & \textbf{27.7}\\
    T & 20.7 & 26.8 & 26.6 & \textbf{39.9}\\
    J & 53.1 & 56.6 & \textbf{64.6} & 57.3\\
    P & 7.9 & 27.7 & \textbf{29.7} & 22.8\\
    \bottomrule
    \end{tabular}
    \caption{The Exact Match (\texttt{EM}) accuracies of the compared ablations on \benchName.}
    \label{tab:ablationEM}
    \end{small}
\end{table}
%% EM

%% NS errors
\begin{figure*}[t]
\small
\centering
\begin{tabular}{p{0.9\textwidth}}
\toprule
\textbf{Question}\\
\midrule
What are the names, countries, and ages for every singer in descending order of age?\\
\midrule
\textbf{Gold Queries}\\
\midrule
\begin{enumerate}[partopsep=0pt,topsep=0pt,parsep=0pt,leftmargin=*]
\item SELECT name, nationality, age FROM singer ORDER BY age DESC
\item SELECT name, citizenship, age FROM singer ORDER BY age DESC
\end{enumerate}\\
\midrule
\textbf{Outputs of T5-3B with Nucleus Sampling}\\
\midrule
\begin{enumerate}[partopsep=0pt,topsep=0pt,parsep=0pt,leftmargin=*]
\item  \textcolor{green!60!black}{SELECT name, nationality, age FROM singer ORDER BY age DESC}
\item  SELECT name, nationality, age FROM singer ORDER BY age DESC
\item  SELECT name, nationality, age FROM singer ORDER BY age DESC
\item  SELECT name, nationality, age FROM singer ORDER BY age DESC
\item  SELECT name, nationality, age FROM singer ORDER BY age DESC
\end{enumerate}\\
\bottomrule
\end{tabular}
\caption{Nucleus Sampling shows virtually no diversity in top-$5$ outputs due to a highly skewed probability distribution leading to the same tokens being sampled each time.}
\label{fig:nserrors}
\end{figure*}

%% Figures not of any appendix section

\begin{figure*}[t]
\centering
\small
\begin{tabular}{p{0.9\textwidth}}
\toprule
\textbf{Example Template Outputs with Beam Search}\\
\midrule
\begin{enumerate}[partopsep=0pt,topsep=0pt,parsep=0pt,leftmargin=*]
\item  SELECT column, AVG(column) FROM table GROUP BY column ORDER BY column
\item  SELECT column, AVG(column) FROM table GROUP BY column ORDER BY column ASC
\item  SELECT column, AVG(column) FROM table GROUP BY column ORDER BY column DESC
\item  SELECT AVG(column), column FROM table GROUP BY column ORDER BY column
\item  SELECT column, AVG(column) FROM table ORDER BY column
\end{enumerate}\\
\bottomrule
\end{tabular}
\caption{Vanilla Beam Search is inadequate to elicit meaningful template diversity. In particular, diversity in the number of \texttt{JOIN}s or selections is lacking.}
\label{fig:templateBSdiversity}
\end{figure*}

We use the directive and prompt shown in Listings~\ref{lst:benchdirective} and \ref{lst:benchprompt} respectively.
The database ID and its schema go into \texttt{[DB\_ID]} and \texttt{[SCHEMA]}, respectively. The question is passed at the end in the placeholder \texttt{[QUESTION]}. Our demonstration for ChatGPT consists of using the question ``\emph{Show the stadium name and the number of concerts in each stadium}'', and the output used for demonstration is shown in Listing~\ref{lst:outputs}.

\begin{lstlisting}[float=*, caption={The demonstrated outputs for the one-shot example with the query ``\emph{Show the stadium name and the number of concerts in each stadium}''.}, label={lst:outputs}, captionpos=b, belowskip=-0.8 \baselineskip]
{
    "queries": [
        "select t2.name, count(*) from concert as t1 join stadium as t2 on t1.stadium_id = t2.stadium_id group by t1.stadium_id",
        "select t3.name, count(*) from concert as t1 join stadium as t2 on t1.stadium_id = t2.stadium_id join singer as t3 on t1.singer_id = t3.singer_id group by t1.stadium_id",
        "select t3.name, count(*) from concert as t1 join stadium as t2 on t1.stadium_id = t2.stadium_id join singer_in_concert as t3 on t1.concert_id = t3.singer_id group by t1.stadium_id",
        "select t3.name, count(*) from concert as t1 join stadium as t2 on t1.stadium_id = t2.stadium_id join singer_in_concert as t3 on t1.concert_id = t3.singer_id group by t1.stadium_id",
        "select t1.name, count(*) from stadium as t1 join concert as t2 on t1.stadium_id = t2.stadium_id group by t1.stadium_id"
    ]
}
\end{lstlisting}

\subsection{Text-to-SQL (OpenAI Codex)}
We found that asking Codex to produce multiple SQLs in the same output did not have the desired effect, as it did not usually conform to the number of outputs or the format. Therefore, we instead prompt Codex multiple times with a temperature of $0.6$ (as recommended by OpenAI to elicit creativity) and a top-$p$ of $0.7$ to get its outputs. To this end, we found both zero and one-shot prompting ineffective in conveying to Codex the specific format of the output (unlike ChatGPT). In contrast, we found that few (specifically, two) shot prompting to work much better, and therefore proceeded with that alternative. Our two demonstrations as well as the query prompt follow the format of Listing~\ref{lst:codexprompt}. The output formatting is simply the SQL query string inside curly braces. The two demonstrated examples are replicated in Listing~\ref{lst:codexex}.

\begin{lstlisting}[float=*, caption={The format of both the demonstrated and query examples}, label={lst:codexprompt}, captionpos=b, belowskip=-0.8 \baselineskip]
# Use the schema links to generate the SQL query for the question

[SCHEMA]

Convert the following English question to SQL queries compatible with the above schema.

Use simply the column name for selections in simple queries. For queries with joins, use t1, t2 and so on as aliases for the tables, and use t1.column, t2.column and so on for the column selections.

Question: [QUESTION]
\end{lstlisting}

\begin{lstlisting}[float=*, caption={The examples used as demonstrations for OpenAI Codex. The ``Question'' and ``Query'' indicators are just for clarity, and the formatting is as per Listing~\ref{lst:codexprompt}.}, label={lst:codexex}, captionpos=b, belowskip=-0.8 \baselineskip]
Question 1: List the official name and status of the city with the largest population.
Query 1: SELECT official_name, status FROM city ORDER BY population DESC LIMIT 1
Question 2: Show the stadium name and the number of concerts in each stadium.
Query 2: SELECT t1.name, count(*) FROM stadium AS t1 JOIN concert AS t2 ON t1.stadium_id = t2.stadium_id GROUP BY t1.stadium_id
\end{lstlisting}

\begin{lstlisting}[float=*, caption={An alternate prompt used by prior work that we tried.}, label={lst:altprompt}, captionpos=b, belowskip=-0.8 \baselineskip]
### Generate 5 possible sqlite SQL queries ending with ';' for the question given in triple backticks, with no explanation.
### Sqlite SQL tables, with their properties: 
#
[DB_SCHEMA]
#
### ``` [QUESTION] ```
\end{lstlisting}

\section{Alternate Prompts with ChatGPT}
\label{ap:altprompt}
Before settling on our choice, we also experimented with existing prompts used by prior work (zero-shot, as opposed to our one-shot method). In particular, we tried the prompt used by \cite{ChatGPTEval} to evaluate ChatGPT on our benchmark with minor modifications (asking for five outputs instead of one). We showcase it in Listing~\ref{lst:altprompt}.\\

However, as shown in \tref{tab:altchatgpt}, the results with this prompting method always lag behind those obtained with our main choice.
\begin{table}[H]
    \begin{small}
    \centering
    \begin{tabular}{c|c|c}
    \toprule
    \makecell{Kind of\\Ambiguity} & \makecell{ChatGPT (\%)\\(Our Prompt)} & \makecell{ChatGPT (\%)\\ \cite{ChatGPTEval}}\\
    \midrule
    C & \textbf{22.7} & 11.2\\
    T & \textbf{37.3} & 15.5\\
    J & \textbf{15.6} & 0.0 \\
    P & \textbf{8.9} & 6.9 \\
    \bottomrule
    \end{tabular}
    \caption{The results of ChatGPT with the prompt of \cite{ChatGPTEval} lags behind those obtained with our prompt. The better numbers are \textbf{bolded}.}
    \label{tab:altchatgpt}
    \end{small}
\end{table}
Therefore, we decided to stick with our choice for the comparison in Subsection~\ref{sub:comp}.
\section{Exact Set Match Accuracies for the System Comparison and Ablation Study}
\label{ap:fullresults}
Here we report the Exact Match (\texttt{EM}) accuracies of our System Comparison and Ablation Study for both the \emph{\Any} and \emph{\All} modes of evaluation.

The System Comparison on \benchName\ in terms of \texttt{EM}, and of the various decoding algorithms, when applied to T5-3B, are shown in \tref{tab:systemsEM}. We observe that Exact Set Match (\texttt{EM}) follows the same trend as Execution Match (\texttt{EXM}) under both headings, once again demonstrating the superior coverage of \algName. 

The results of our Ablation Study, in turn, are shown in \tref{tab:ablationEM}. The trend of \texttt{EM} also matches that of \texttt{EXM} here.

\section{Inadequacy of Conventional Decoding Algorithms}
\label{ap:decodeinadequate}
In this section, we give some anecdotes to highlight the shortcomings of conventional decoding algorithms for our purposes. The example for the case of Beam Search when used with a Text-to-SQL model was given in the main material as \fref{fig:bserrors}. We also give here an anecdote of Nucleus Sampling in \fref{fig:nserrors}. Strikingly, all the outputs of Nucleus Sampling are the same. This was the case for many of the examples we manually appraised. Upon further investigation, we discovered that the model produced extremely skewed probability distributions for its tokens --- it was not uncommon for certain tokens to be assigned greater than 0.99 probability. This renders conventional decoding algorithms, including sampling-based methods, ineffective. Similarly, we found Beam Search (as well as sampling approaches) to be suboptimal for the case of Text-to-Template conversion, as \fref{fig:templateBSdiversity} exemplifies.

\section{Examples of Templates}
\label{ap:templateinfo}
A template is generated by abstracting away column names, table names, integer constants, and string literals from an SQL query. While these are only a small fraction of the various features of the SQL query, they represent a disproportionately large percentage of viable alternatives - for instance, a column name may be replaced by any of the numerous other ones to generate an (otherwise useless) alternative. By abstracting away these details, we avoid generating spurious alternatives by swapping these features with other ones at the template generation stage. In addition, by generating, e.g., \texttt{column} instead of \texttt{t1.column} for \texttt{t1.name}, we avoid trivial alias swaps. Some examples of templates for a few SQL queries are shown in \tref{tab:tempex}, and the replacements carried out for each kind of abstraction are outlined in \tref{tab:templaterep}.

%% Template examples
\begin{table}[H]
    \begin{small}
    \centering
    \begin{tabular}{l|l}
    \toprule
    \multicolumn{1}{c|}{SQL Query} & \multicolumn{1}{c}{Template}\\ 
    \midrule
    \texttt{SELECT name FROM singer} & \texttt{\makecell[l]{SELECT column FROM\\table}}\\
    \hline
    \texttt{\makecell[l]{SELECT t1.born_state,\\AVG(t2.age) FROM head\\AS t1 JOIN employee\\AS t2 ON t1.emp\_id\\= t2.emp\_id}} & \texttt{\makecell[l]{SELECT column, AVG(\\column) FROM table AS\\t1 JOIN table AS t2\\ON column = column}}\\
    \hline
    \texttt{\makecell[l]{SELECT last\_name FROM\\head WHERE age > 56\\AND first\_name = ``John''}} & \texttt{\makecell[l]{SELECT column FROM\\table WHERE column >\\number AND column =\\string}}\\
    \bottomrule
    \end{tabular}
    \caption{Examples of templates.}
    \label{tab:tempex}
    \end{small}
\end{table}

%% Template replacements
\begin{table}[H]
    \begin{small}
    \centering
    \begin{tabular}{l|l|l}
    \toprule
    Token type & Example & Abstraction\\ 
    \midrule
    Column name & \texttt{t1.name, age} & \texttt{column}\\
    Table name & \texttt{singer} & \texttt{table}\\
    Number & 2,3.5 & \texttt{number}\\
    String & ``California'' & \texttt{string}\\
    \bottomrule
    \end{tabular}
    \caption{The abstractions in a template.}
    \label{tab:templaterep}
    \end{small}
\end{table}

\section{Example Outputs From the Systems}
\label{ap:examples}
We showcase example outputs from three chosen systems - our method, ChatGPT, and T5-3B on the various kinds of ambiguities of \benchName in Figures~\ref{fig:C-example} through \ref{fig:P-example}. Note that the first two outputs of our approach are from T5-3B. 
We observe that our approach is more consistent than the other two in incorporating all the possible queries. 

%% Begin C-example
\begin{figure*}[t]
\small
\centering
\begin{tabular}{p{0.9\textwidth}}
\toprule
\multicolumn{1}{c}{{\Large \textbf{(C)olumn Synonyms}}}\\
\midrule
\textbf{Question}\\
\midrule
What are the names of documents that use templates with the code BK?\\
\midrule
\textbf{Gold Queries}\\
\midrule
\begin{enumerate}[partopsep=0pt,topsep=0pt,parsep=0pt,leftmargin=*]
\item SELECT t2.file\_name FROM templates AS t1 JOIN documents AS t2 ON t1.template\_id = t2.template\_id WHERE t1.template\_type\_code = ``BK''
\item SELECT t2.record\_name FROM templates AS t1 JOIN documents AS t2 ON t1.template\_id = t2.template\_id WHERE t1.template\_type\_code = ``BK''
\end{enumerate}\\
\midrule
\textbf{\algName}\\
\midrule
\begin{enumerate}[partopsep=0pt,topsep=0pt,parsep=0pt,leftmargin=*]
\item \textcolor{green!60!black}{SELECT t1.file\_name FROM documents AS t1 JOIN templates AS t2 ON t1.template\_id = t2.template\_id WHERE t2.template\_type\_code = ``BK''}
\item SELECT t1.file\_name, t1.document\_description FROM documents AS t1 JOIN templates AS t2 ON t1.template\_id = t2.template\_id WHERE t2.template\_type\_code = ``BK''
\item \textcolor{green!60!black}{SELECT t1.record\_name FROM documents AS t1 JOIN templates AS t2 ON t1.template\_id = t2.template\_id WHERE t2.template\_type\_code = ``BK''}
\item SELECT t1.file\_name FROM documents AS t1 JOIN templates AS t2 ON t1.template\_id = t2.template\_id WHERE t2.template\_type\_code = ``BK'' UNION SELECT t1.file\_name FROM documents AS t1 JOIN templates AS t2 ON t1.template\_id = t2.template\_id WHERE t2.template\_type\_code = ``BK''
\item SELECT t1.file\_name FROM documents AS t1 JOIN documents AS t2 ON t1.template\_id = t2.template\_id WHERE t2.template\_type\_code = ``BK''
\end{enumerate}\\
\midrule
\textbf{ChatGPT}\\
\midrule
\begin{enumerate}[partopsep=0pt,topsep=0pt,parsep=0pt,leftmargin=*]
\item SELECT t1.record\_name FROM templates AS t1 JOIN ref\_template\_types AS t2 ON t1.template\_type\_code = t2.template\_type\_code JOIN documents AS t3 ON t1.template\_id = t3.template\_id WHERE t2.template\_type\_code = `BK'
\item SELECT t3.record\_name FROM documents AS t3 JOIN templates AS t2 ON t3.template\_id = t2.template\_id JOIN ref\_template\_types AS t1 ON t1.template\_type\_code = t2.template\_type\_code WHERE t1.template\_type\_code = `BK'
\item SELECT t1.record\_name FROM templates AS t1 JOIN documents AS t2 ON t1.template\_id = t2.template\_id JOIN ref\_template\_types AS t3 ON t1.template\_type\_code = t3.template\_type\_code WHERE t3.template\_type\_code = `BK'
\item SELECT t3.record\_name FROM documents AS t3 JOIN templates AS t2 ON t3.template\_id = t2.template\_id WHERE t2.template\_type\_code = `BK'
\item SELECT t1.record\_name FROM templates AS t1 JOIN documents AS t2 ON t1.template\_id = t2.template\_id WHERE t1.template\_type\_code = `BK'
\end{enumerate}\\
\midrule
\textbf{T5-3B}\\
\midrule
\begin{enumerate}[partopsep=0pt,topsep=0pt,parsep=0pt,leftmargin=*]
\item \textcolor{green!60!black}{SELECT t1.file\_name FROM documents AS t1 JOIN templates AS t2 ON t1.template\_id = t2.template\_id WHERE t2.template\_type\_code = ``BK''}
\item SELECT t1.file\_name, t1.document\_description FROM documents AS t1 JOIN templates AS t2 ON t1.template\_id = t2.template\_id WHERE t2.template\_type\_code = ``BK''
\item SELECT file\_name FROM documents AS t1 JOIN templates AS t2 ON t1.template\_id = t2.template\_id WHERE t2.template\_type\_code = ``BK''
\item SELECT t1.file\_name, t1.record\_name FROM documents AS t1 JOIN templates AS t2 ON t1.template\_id = t2.template\_id WHERE t2.template\_type\_code = ``BK''
\item SELECT t1.document\_name FROM documents AS t1 JOIN templates AS t2 ON t1.template\_id = t2.template\_id WHERE t2.template\_type\_code = ``BK''
\end{enumerate}\\
\bottomrule
\end{tabular}
\caption{Example outputs from our system, ChatGPT and T5-3B on the C portion of \benchName}
\label{fig:C-example}
\end{figure*}
%% End C-example

%% Begin T-example
\begin{figure*}[t]
\small
\centering
\begin{tabular}{p{0.9\textwidth}}
\toprule
\multicolumn{1}{c}{{\Large \textbf{(T)able Synonyms}}}\\
\midrule
\textbf{Question}\\
\midrule
Show the name and the release year of the song by the youngest singer.\\
\midrule
\textbf{Gold Queries}\\
\midrule
\begin{enumerate}[partopsep=0pt,topsep=0pt,parsep=0pt,leftmargin=*]
\item SELECT song\_name, song\_release\_year FROM artist ORDER BY age LIMIT 1
\item SELECT song\_name, song\_release\_year FROM performer ORDER BY age LIMIT 1
\end{enumerate}\\
\midrule
\textbf{\algName}\\
\midrule
\begin{enumerate}[partopsep=0pt,topsep=0pt,parsep=0pt,leftmargin=*]
\item \textcolor{green!60!black}{SELECT song\_name, song\_release\_year FROM artist ORDER BY age LIMIT 1}
\item SELECT t2.song\_name, t2.song\_release\_year FROM artist AS t1 JOIN song\_name AS t2 ON t1.song\_name = t2.song\_name ORDER BY t1.age LIMIT 1
\item \textcolor{green!60!black}{SELECT song\_name, song\_release\_year FROM performer ORDER BY age LIMIT 1}
\item SELECT song\_name, song\_release\_year FROM performer ORDER BY age asc LIMIT 1
\item SELECT song\_name, song\_release\_year FROM concert ORDER BY age LIMIT 1
\end{enumerate}\\
\midrule
\textbf{ChatGPT}\\
\midrule
\begin{enumerate}[partopsep=0pt,topsep=0pt,parsep=0pt,leftmargin=*]
\item \textcolor{green!60!black}{SELECT name, song\_release\_year FROM performer WHERE age = (SELECT MIN(age) FROM performer) LIMIT 1}
\item \textcolor{green!60!black}{SELECT name, song\_release\_year FROM artist WHERE age = (SELECT MIN(age) FROM artist) LIMIT 1}
\item SELECT t1.name, t1.song\_release\_year FROM performer AS t1 JOIN singer\_in\_concert AS t2 ON  t1.singer\_id = t2.singer\_id WHERE t1.age = (SELECT MIN(age) FROM performer) LIMIT 1
\item SELECT t1.name, t1.song\_release\_year FROM artist AS t1 JOIN singer\_in\_concert AS t2 ON  t1.singer\_id = t2.singer\_id WHERE t1.age = (SELECT MIN(age) FROM artist) LIMIT 1
\item SELECT t1.name, t1.song\_release\_year FROM performer AS t1 JOIN singer\_in\_concert AS t2 ON t1.singer\_id = t2.singer\_id WHERE t1.age = (SELECT MIN(age) FROM (SELECT * FROM performer UNION SELECT * FROM artist)) LIMIT 1
\end{enumerate}\\
\midrule
\textbf{T5-3B}\\
\midrule
\begin{enumerate}[partopsep=0pt,topsep=0pt,parsep=0pt,leftmargin=*]
\item \textcolor{green!60!black}{SELECT song\_name, song\_release\_year FROM artist ORDER BY age LIMIT 1}
\item SELECT t2.song\_name, t2.song\_release\_year FROM artist AS t1 JOIN song\_name AS t2 ON t1.song\_name = t2.song\_name ORDER BY t1.age LIMIT 1
\item SELECT t2.song\_name, t2.song\_release\_year FROM artist AS t1 JOIN song\_name AS t2 ON t1.song\_name = t2.song\_name WHERE t1.age = (SELECT MIN(age) FROM artist)
\item SELECT song\_name, song\_release\_year FROM artist ORDER BY age asc LIMIT 1
\item SELECT t2.song\_name, t2.song\_release\_year FROM artist AS t1 JOIN performer AS t2 ON t1.singer\_id = t2.singer\_id ORDER BY t1.age LIMIT 1
\end{enumerate}\\
\bottomrule
\end{tabular}
\caption{Example outputs from our system, ChatGPT and T5-3B on the T portion of \benchName}
\label{fig:T-example}
\end{figure*}
%% End T-example

%% Begin J-example
\begin{figure*}[t]
\small
\centering
\begin{tabular}{p{0.9\textwidth}}
\toprule
\multicolumn{1}{c}{{\Large \textbf{(J)oin Ambiguity}}}\\
\midrule
\textbf{Question}\\
\midrule
Return the name, location, and district of all shops in descending order of number of products.\\
\midrule
\textbf{Gold Queries}\\
\midrule
\begin{enumerate}[partopsep=0pt,topsep=0pt,parsep=0pt,leftmargin=*]
\item SELECT name, location, district FROM shop ORDER BY number\_products DESC
\item SELECT t1.name, t1.location, t2.district FROM shop AS t1 JOIN shop\_district AS t2 ON t1.shop\_id = t2.shop\_id ORDER BY t1.number\_products DESC
\end{enumerate}\\
\midrule
\textbf{\algName}\\
\midrule
\begin{enumerate}[partopsep=0pt,topsep=0pt,parsep=0pt,leftmargin=*]
\item \textcolor{green!60!black}{SELECT name, location, district FROM shop ORDER BY number\_products DESC}
\item SELECT name, location and district FROM shop ORDER BY number\_products DESC
\item SELECT name, district, location FROM shop ORDER BY number\_products DESC
\item SELECT name, district FROM shop ORDER BY number\_products DESC
\item \textcolor{green!60!black}{SELECT t1.name, t1.location, t2.district FROM shop AS t1 JOIN shop\_district AS t2 ON t1.shop\_id = t2.shop\_id ORDER BY t1.number\_products DESC}
\end{enumerate}\\
\midrule
\textbf{ChatGPT}\\
\midrule
\begin{enumerate}[partopsep=0pt,topsep=0pt,parsep=0pt,leftmargin=*]
\item SELECT name, location, district FROM shop JOIN shop\_district ON shop.shop\_id = shop\_district.shop\_id ORDER BY number\_products DESC
\item SELECT name, location, district FROM shop\_district JOIN shop ON shop.shop\_id = shop\_district.shop\_id ORDER BY number\_products DESC
\item SELECT s.name, s.location, sd.district FROM shop AS s JOIN shop\_district AS sd ON s.shop\_id = sd.shop\_id JOIN shop AS s2 ON s2.shop\_id = sd.shop\_id GROUP BY s.name, sd.district, s.location ORDER BY sum(s2.number\_products) DESC
\item SELECT s.name, s.location, sd.district FROM shop\_district AS sd JOIN shop AS s ON s.shop\_id = sd.shop\_id JOIN shop AS s2 ON s2.shop\_id = sd.shop\_id GROUP BY s.name, sd.district, s.location ORDER BY sum(s2.number\_products) DESC
\item SELECT name, location, district FROM shop\_district JOIN shop ON shop.shop\_id = shop\_district.shop\_id ORDER BY (SELECT count(*) FROM employee WHERE employee.city = shop.location) DESC
\end{enumerate}\\
\midrule
\textbf{T5-3B}\\
\midrule
\begin{enumerate}[partopsep=0pt,topsep=0pt,parsep=0pt,leftmargin=*]
\item  \textcolor{green!60!black}{SELECT name, location, district FROM shop ORDER BY number\_products DESC}
\item SELECT name, location and district FROM shop ORDER BY number\_products DESC
\item SELECT name, location, district FROM shop ORDER BY number\_products DESC
\item SELECT name, location, district form shop ORDER BY number\_products DESC
\item SELECT name, location, district of shop ORDER BY number\_products DESC
\end{enumerate}\\
\bottomrule
\end{tabular}
\caption{Example outputs from our system, ChatGPT and T5-3B on the J portion of \benchName}
\label{fig:J-example}
\end{figure*}
%% End J-example

%% Begin P-example
\begin{figure*}[t]
\small
\centering
\begin{tabular}{p{0.9\textwidth}}
\toprule
\multicolumn{1}{c}{{\Large \textbf{(P)re-Computed Aggregates}}}\\
\midrule
\textbf{Question}\\
\midrule
What is the average and the maximum capacity of all stadiums?\\
\midrule
\textbf{Gold Queries}\\
\midrule
\begin{enumerate}[partopsep=0pt,topsep=0pt,parsep=0pt,leftmargin=*]
\item SELECT AVG(capacity), MAX(capacity) FROM stadium
\item SELECT avg\_capacity, max\_capacity FROM stadium\_capacity
\end{enumerate}\\
\midrule
\textbf{\algName}\\
\midrule
\begin{enumerate}[partopsep=0pt,topsep=0pt,parsep=0pt,leftmargin=*]
\item \textcolor{green!60!black}{SELECT avg\_capacity, max\_capacity FROM stadium\_capacity}
\item SELECT avg\_capacity (SELECT AVG(capacity) FROM stadium\_capacity) and max\_capacity (SELECT MAX(capacity) FROM stadium)
\item \textcolor{green!60!black}{SELECT AVG(capacity), MAX(capacity) FROM stadium}
\item SELECT AVG(capacity), MAX(max\_capacity) FROM stadium\_capacity
\item SELECT AVG(capacity), MAX(capacity), stadium\_id FROM stadium\_capacity
\end{enumerate}\\
\midrule
\textbf{ChatGPT}\\
\midrule
\begin{enumerate}[partopsep=0pt,topsep=0pt,parsep=0pt,leftmargin=*]
\item \textcolor{green!60!black}{SELECT AVG(capacity), MAX(capacity) FROM stadium}
\item SELECT AVG(t1.capacity), MAX(t1.capacity) FROM stadium AS t1
\item SELECT AVG(t1.capacity), t1.max\_capacity FROM stadium\_capacity AS t1
\item SELECT AVG(t1.capacity), MAX(t2.capacity) FROM stadium AS t1 JOIN stadium AS t2 on t1.capacity $<$= t2.capacity group by t1.capacity
\item SELECT AVG(t2.capacity), MAX(t2.capacity) FROM concert AS t1 JOIN stadium AS t2 on t1.stadium\_id = t2.stadium\_id
\end{enumerate}\\
\midrule
\textbf{T5-3B}\\
\midrule
\begin{enumerate}[partopsep=0pt,topsep=0pt,parsep=0pt,leftmargin=*]
\item \textcolor{green!60!black}{SELECT avg\_capacity, max\_capacity FROM stadium\_capacity}
\item SELECT avg\_capacity (SELECT AVG(capacity) FROM stadium\_capacity) and max\_capacity (SELECT MAX(capacity) FROM stadium)
\item SELECT avg\_capacity (SELECT AVG(capacity) FROM stadium\_capacity) and max\_capacity (SELECT max\_capacity FROM stadium\_capacity) FROM stadium
\item SELECT AVG(capacity), MAX(capacity) FROM stadium\_capacity
\item SELECT avg\_capacity (SELECT AVG(capacity) FROM stadium\_capacity) and max\_capacity (SELECT MAX(capacity) FROM stadium\_capacity)
\end{enumerate}\\
\bottomrule
\end{tabular}
\caption{Example outputs from our system, ChatGPT and T5-3B on the P portion of \benchName}
\label{fig:P-example}
\end{figure*}
%% End P-example

\end{document}